\documentclass{llncs}

\usepackage{amsmath}
\usepackage{graphicx}
\usepackage{color}
\usepackage{url}
\usepackage{cite}
\usepackage{booktabs} 
\usepackage{subfig}
\usepackage{placeins}
\usepackage{algorithm}
\usepackage[noend]{algpseudocode}

\usepackage{amsmath}

\DeclareGraphicsExtensions{.pdf,.png,.jpg}

\usepackage[english]{babel}
\usepackage{amssymb}
\usepackage{amsmath}
\usepackage{graphicx}
\usepackage{dcolumn}
\usepackage{hyperref}

\begin{document}

\frontmatter          
\pagestyle{empty}  

\title{Towards a learning-based performance modeling for accelerating Deep Neural Networks}
 \author{Damiano Perri\inst{1}$^{ORCID:0000-0001-6815-6659}$ \and Paolo Sylos Labini\inst{2} \and Osvaldo Gervasi\inst{1}$^{ORCID: 0000-0003-4327-520X}$ \and Sergio Tasso\inst{1}$^{ORCID: 0000-0001-9174-9065}$ \and Flavio Vella\inst{2}$^{ORCID: 0000-0002-5676-9228}$ 
 }
%

%
\institute{University of Perugia, Dept. of Mathematics and Computer Science,\\
Perugia, Italy\\
\and Lab for Advanced Computing and Systems , Free University of Bozen-Bolzano,\\ Bolzano, Italy\\
}

\maketitle
\begin{abstract}
Emerging applications such as Deep Learning are often data-driven, thus traditional approaches based on auto-tuners are not performance effective across the wide range of inputs used in practice. 
In the present paper, we start an investigation of predictive models based on machine learning techniques in order to optimize Convolution Neural Networks (CNNs). 
As a use-case, we focus on the ARM Compute Library which provides three different implementations of the convolution operator at different numeric precision. Starting from a collation of benchmarks, we build and validate models learned by Decision Tree and naive Bayesian classifier. 
Preliminary experiments on Midgard-based ARM Mali GPU show that our predictive model outperforms all the convolution operators manually selected by the library.  
\end{abstract}

\section{Introduction}
With the advent of big-data and data-driven applications such as deep learning, convolutional neural network for image classification and graph analytics among the others, the traditional library design looses performance portability mainly due to the unpredictable size and structure of the data. 
Specific algorithms and implementations are mostly designed by taking into account specific characteristics of the input or the targeting architecture.  
Autotuners partially mitigate this problem by adapting the implementation to the underline architecture, for example by selecting the best Local Work Size on OpenCL compliant GPUs\cite{CLTune2015}.
Vendors libraries (e.g., Nvidia CuBLAS) still apply manual heuristics in
order to select at runtime highly-optimized code for specific inputs. 
Convolution, the most crucial and computationally expensive part of both the training and inference step in CNNs, represents a notable example where it is quite hard to determine the best implementation for a given input\cite{krizhevsky2012imagenet}. The choice among direct, Image-to-column, FFT-based or Winograd-based algorithms may vary even in the same CNN, since different layers requires convolution operators to act on different input sizes. 

The aim of this work is to study a model-driven approach in order to improve the performance of the ARM Compute library by predicting the best convolution methods for a given convolution layers. 
Therefore, the model must be able to discriminate the architecture, numerical precision and input size.  

The contributions of this preliminary work is twofold:
\begin{itemize}
    \item we describe a methodology to generate the dataset used to build a predictive model.
    \item we evaluate a machine-learning based model on a convolutional neural networks on ARM GPUs
\end{itemize}
The rest of paper is organized as follows. 
Section \ref{sec:related} provides a brief description of related works.
The background is given in Section \ref{background}.
The main contribution of the work is reported on Section \ref{method} (Methodology)and Section \ref{sec:exp} (Experiments).
Section \ref{sec:conclusion} concludes the paper by underlining the lesson learned and the possibilities for future works. 

\section{Related Work}
\label{sec:related}
The size of the input matrices and kernels has been carefully analyzed as a function of different systems (server CPU, server GPU, mobile phone) by M. Cho and D. Brand in \cite{art4}. The authors carried out a systematic comparison between the main convolution methods (Winograd, image to column and Generic Matrix Multiplication and Fast Fourier Trasform). 
The effects of multiple input channels have been studied by A. Vasudevan et al. in \cite{art5}, who carried out a systematic analysis of the performances of the Image to Column and GEMM method varying the input channels and kernel sizes (e.g: 3x3 and 5x5), benchmarking the performances on various architectures (Intel Core i5-4570 and ARM Cortex-A57) on different neural networks (VGG-16, GoogleNet, AlexNet).
The effects on performances of varying the accuracy has been studied by Vella et al.~\cite{lokhmotov2018multi}. 
As reported, a reduction of the accuracy by 7\% increases three times the performances on a Firefly board. 
Input aware techniques~\cite{falch2015machine} are recently used to address the problem of performance portability on different applications~\cite{hou2017auto, cosenza2017autotuning}.
Other seminal works successfully investigated predictive models for the performance modeling\cite{singer2000learning} for accelerating linear algebra routines\cite{cianfriglia2018model} or improving the scheduling of processes on hybrid systems\cite{tasso2015simulation}.
Their results inspired us to adopt a Machine Learning approach to find optimal implementations of the convolution operation and provided several insights for our experimental setup.

\section{Background\label{background}}
Convolutional Neural Networks (CNNs) are a class of deep, feed-forward artificial neural networks that are often used to recognize objects in images.
CNNs are composed by a set of layers linked by consecutive convolution operations.
In image classification tasks, each layer is organized as a multi-channel, two-dimensional collections of neurons. Layers, along with the convolution operators acting on them, are characterized by several parameters such as width, height, depth, filter dimension, pad and stride. The last layer (output layer) is reduced to a single vector of probability scores, so that a CNN transforms the original pixel values of an input image to the final class scores.
The shape of such layers is usually fixed beforehand, while the kernel coefficient of their convolution operators are tuned through training. 

\subsubsection{Convolution}
A great number of standard signal-processing operations are described by linear and time invariant operators. Their action on a function $f$ can be implemented through convolution with a filter (or kernel) $k$, indicated with the $*$ operator and defined as:

\begin{equation}
(f*k)(t)=\int f(\tau)k(t-\tau)d\tau  \label{imgconv}
\end{equation}

Often, convolution is applied to discrete, finite signals, such as digital images. 
For computing a convolution in the notable case of 2-dimensional, single channel digital images, a variation of \eqref{imgconv} is employed: 

\begin{equation}
(f*k)(x,y)=\sum_{i=0}^M\sum_{j=0}^N f(i,j)k(x-i,y-j) \label{dirconv}
\end{equation}

Thus, convolution changes the value (color, transparency, etc.) of a pixel to a weighted sum of all other pixels. These weights are the entries of the kernel matrix $k$, translated so that its center lies on the target pixel.
Usually, the kernel matrix is null everywhere but a small region around its center, so that the value of a point after a convolution depends only on its close neighbours. The size of the non-zero part of the kernel may be arbitrary, but a $3x3$ matrix if often used in image processing applications.

Convolution is thus a general purpose filter effect for images. Varying the convolution kernel, we may obtain a variety of effects, such as enhancing the edges, increasing the contrast, dilating or eroding the area occupied by the objects in the picture, and so on. 

Performing a "direct convolution" and computing directly \eqref{dirconv} can be unnecessarily costly, so other indirect methods are often preferred. The following sections reports a brief overview of some of these methods.

\subsubsection{Coppersmith Winograd}
In 1969, Strassen developed a matrix multiplication algorithm with complexity \(O(n^{2.81})\), outperforming the standard \(O(n^{3})\) algorithm through the use of a number of intermediate products and additions.
Subsequently, in 1986, he developed the "laser method", which further reduced the complexity to \(O(n^{2.48})\).
The following year, Coppersmith and Winograd developed a faster, now popular algorithm with complexity equal to \(O(n^{2.375477})\).  
Although this algorithm comes with a lower asymptotic cost than its predecessors, a large multiplicative constant in the omega notation makes it truly efficient only when the matrices have particularly large dimensions. 
Since it is possible to recast \eqref{dirconv} as a product of matrices, a Winograd-based convolution is possible, and actually very efficient and numerically stable, especially for small 3x3 kernels.

\subsubsection{Fast Fourier Transform}
Since Fourier functions are the eigenfunctions of the convolution operator, convolution is easily performed in the Fourier domain as a multiplication between the function and the kernel coefficients. 
The Fast Fourier Transform (FFT) is a computationally fast way of obtaining the Fourier coefficients of a signal. A convolution through FFT can be very efficient when it involves large filters, since the cost of applying the filter in the Fourier domain is small compared to that of transforming the two signals. 
Unfortunately, as already noted, most modern applications use very small filters that can easily run on highly parallel system, making FFT-based convolution less convenient. Combined with an inherently weak numerical precision, this makes it hard for the FFT procedure to compete with some other methods. 
For comparison, we report here results extracted from the work of Andrew Lavin and Scott Gray on the comparison between FFT-based and Winograd-based convolution methods. In November 2015, they tested the two algorithms by running them on nVidia Maxwell architecture, specifically using a Titan X graphics card.

Using 32 bit floating point and a 3x3 filter, Winograd performed better: it achieved an error rate of \(1.53*10^{-5}\), against the \( 4.01*10^{-5}\) of the FFT. 
Interestingly, when using 16 bit floating point, the two techniques obtained the same level of precision, but in both cases the Winograd speed performances were better by a factor of 2.44.

\subsubsection{Image to column and GEMM}
GEneral Matrix to Matrix Multiplication (GEMM) indicates the standard low-level routine for performing matrix-matrix multiplications. Its implementations are usually extremely optimized for speed and can benefit from special floating point hardware. 
Since images and kernels are represented in memory as five-dimensional arrays (colored RGB), it is necessary to reshape them into 2D matrices in order to perform GEMM.
To this end, a color channel is first selected, and then the an \textit{Image to column} procedure is applied.
Image-to-column rearranges discrete image blocks into columns, and then dispose the concatenated columns in a new matrix.
The order of the columns in the new matrix is determined by traversing the original image in a column-wise manner.
This operation has the considerable disadvantage of increasing the occupied memory, since the pixels of the image are replicated for the generation of the new matrix. Once the matrix associated with the image is obtained, the same procedure is applied to the kernel array, and a new matrix is then generated to be used for multiplication.
Since GEMM is highly optimized, it can allow better performance than direct convolution.
As can be understood, the occupation of memory increases as a result of the generation of several new matrices: with a \textit{n*n} kernel matrix, for example, a column matrix which is \(k^{2}\) times larger than the original image is generated. \\
Yet, in most situations this consistent memory cost comes with a yet more consistent speedup, so that the \textit{image to column and GEMM} approach is employed by a number of Deep Neural Network (DNN) frameworks that target GPUs such as Caffe, Theano and Torch.

\subsection{Supervised Classifiers} 
In the present work we evaluate two of the simplest supervised machine learning methods used for classification and regression, the Decision Tree (DT)~\cite{maron1961automatic}, and naive Bayesian classifier (nBC)~\cite{safavian1991survey}. These straightforward, white-box models greatly simplify this preliminary study and allow us to concentrate on the feasibility of the proposed task.  \\
In the future, we plan to investigate the relation between the characteristics of the input and the parallel implementation of convolution operator provided by the ARM Compute Library through the use of more sophisticated classifiers. 

\subsubsection{Decision Tree}
A \textit{decision tree} is an abstract structure similar to a flow chart graph. In such a tree, nodes identifies decision points, or tests, whit arcs representing outcomes of such test. When the task is classification, leafs are interpreted as class labels, so that traversing the tree from the root to a leaf determines a solution of the decision problem.  
When creating a decision tree for a particular classification problem, one aims to minimize the overall depth while maximizing accuracy, so that the average classification instance is solved traversing the smallest possible number of decision nodes. 
We employed standard ML tools like pruning and used metrics such as the Gini index to reduce complexity and limit overfitting in our DTs.


\subsubsection{Naive Bayesian Classifier}
A Bayesian network is described by a direct acyclic graph, with nodes representing random variables and arcs representing dependencies. Linked nodes shares a direct dependency, while unconnected ones are implied to be conditionally independent. 
In such networks, nodes have a probability distribution that can be assumed or calculated from their neighbours'. This makes it very easy to ask and answer queries about the probability of a variable given some evidence on the others. 
In a naive Bayes classifier, all class labels are considered conditionally independent from each other, and depends only on a single input parent node. Despite their extreme simplification, nBCs have demonstrated exceptionally capable in a variety of real classification tasks. 
The distributions and the dependency structure of a Bayesian network can be constructed ad-hoc from previous knowledge of the system or learned from a dataset, and a variety of algorithms exists for training naive Bayes classifiers. Our choices in this regard are described in the next section along with the employed methodology. 

\section{Methodology and framework description}\label{method}
The proposed methodology can be logically divided in three steps: the \textit{dataset generation}, where we studied the performances of three convolution implementations on a variety of CNN layer shapes, recording the most successful in each instance; the \textit{training}, where we used the dataset to train our classifiers at coupling a layer shape with the optimal convolution implementation for that layer; \\
and finally the \textit{model validation}, were we investigated the performance of our model-driven convolution against the optimal and standard approaches.\\

In what follows, we describe these steps and provide some information on the details of our implementation. The code is available git on \href{https://github.com/DamianoP/AdaptiveMethods}.

\subsubsection{Dataset}
In the first part, we evaluated the performance of the direct, Winograd-based and GEMM-based implementations of the convolution operator. We generated the dataset by collecting the outcome of more than 4000 experiments on artificial CNN layer architectures, stored the performances of each implementation on each layer and identified the fastest in view of the training step. \\

Each convolution layer, as mentioned in Section~\ref{background}, is completely described by five parameters. Three regards the input image: its width $W$, its height $H$, its number of channels $C\_IN$. 
Two characterize the kernel: its side length $KERNEL\_SIZE$, and the number of output channels $C\_OUT$.
We generate the layers varying each parameter separately. Specifically, the parameter \texttt{W} and \texttt{H} can take the values 7, 128 or 256. 
The parameter \texttt{C\_IN} ranges between 3 and 2048 with a multiplicative factor of 32. 384 and 768 were also added to these values.
The parameter \texttt{KERNEL\_SIZE} ranges between 1 and 11 with an increment factor of 1.
The parameter \texttt{C\_OUT} ranges between 8 and 1024 with a multiplicative factor of 2, 384 and 768 were also added to these values. The stride and padding parameters are set to 1.

\begin{figure}[!htb]
	\begin{center}
		\includegraphics[width=1\textwidth]{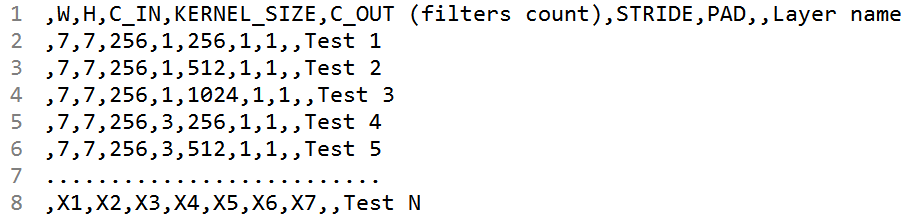}
		\caption{Example of tensor shapes.\label{fig:csv}}
	\end{center}
\end{figure}

Our python script \texttt{tool-prepare-dataset} generates a dataset of performances by executing NNTest over several artificial CNN shapes.
The dataset is stored in a \texttt{.csv} as list of tuples, each containing the feature set of the layer a label with the fastest implementation. An example is reported in Figure
\ref{fig:csv}.

After generation, each input shape was used to evaluate the three implementations provided by the ARM Compute Library and by CK NNTest: direct convolution, winograd, Image to column and GEMM.
This step returns two different files as output. 
The first one is used by existing ML framework (\texttt{.arff} file) like Weka\cite{hall2009weka} and Scikit-learn\cite{Pedregosa:2011:SML:1953048.2078195}.
The second one represent the dataset. Each row is a pair (tensor, label).
Specifically the label is an ordered list of pairs (algorithm name, execution time).

\subsubsection{Training}
In this phase, we trained the Bayesian and the decision tree classifiers on the dataset. The training, in our case carried through the Scikit-Learn python library, aimed to predict performances of convolution implementation based on the feature set of a layer. 

In our code, the pyhton script \texttt{modelGenerator.py},
takes an \texttt{.arff} dataset file as input and outputs a \texttt{.joblib} file that contains the trained model.
Based on the information contained in the \texttt{.arff} and \texttt{.csv} file from the dataset generation phase, we derived the optimal classifier parameter to be used in the next phase for the model evaluation.

\subsubsection{Model evaluation}
Finally, we evaluate the quality of the model in terms of accuracy and performance. We test our classifiers on two real-world CNNs: Inception v3\cite{szegedy2016rethinking}, composed by 66 convolution layers, and MobileNets, composed by 15 convolution layers. 

In our code, for each network we initialize a new classifier, loading in memory the data of the previously trained \texttt{.joblib} model. 
Our script scans the \texttt{.csv} layers list and follows this procedure:
\begin{itemize}
	\item The classifier predicts the fastest algorithm for the current layer.
	\item The script retrieves the optimal implementation using the ranking file, for comparison with the classifier's choice. 	
	\item The script stores the calculation times of the three implementation, and the classifier prediction.
\end{itemize}

Finally, a summary is generated, storing the calculation time of the entire network. The results are plotted in a figure such as \ref{fig:mobilenetsGBBT}, the details of which will be explained in the next section.  


\section{Preliminary Results}
\label{sec:exp}
Before discussing preliminary results, we describe the hardware/software infrastructure used for the experiment below.  
\subsection{Experimental setup}
The hardware setup used for the tests is an ARM Soc with  an ARM Mali-T860 equipped with 4 Mali core able to operate at 2GHz of frequency and 4 GB of DDR3.

We used the ARM Compute Library, an open-source collection of low-level routines optimized for ARM CPU and GPU architectures targeted at image processing, computer vision, and machine learning.
It provides basic arithmetic, mathematical, and binary operators and CNN building blocks. As for convolution, it is implemented in three different ways: image to column and GEMM, direct convolution and Winograd. 
All those methods can be selected at run-time. Depending on the ARM architecture, numeric precision and specific input each methods can exhibit different performance\cite{lokhmotov2018multi,zheng2018optimizing}. 

For benchmarks of each convolution implementation and for the generation of the datasets we used the NNTest library\cite{lokhmotov2018multi, fursin2010collective}, an open-source library for collaboratively validating, benchmarking and optimizing neural net operators across platforms, frameworks and datasets. 

Concerning the model generation frameworks, we use Weka and Scikit-learn. They provide several classification, regression and clustering algorithms. This was used for the training, the tuning and the validation of predictive models.

\subsection{Results}
\begin{figure}
  \centering
  \caption{Inception V3: naive Bayes vs Decision Tree}

  \subfloat[All convolution layers naive Bayes]{\label{fig:inceptionGBBT}\includegraphics[width=0.55\textwidth]{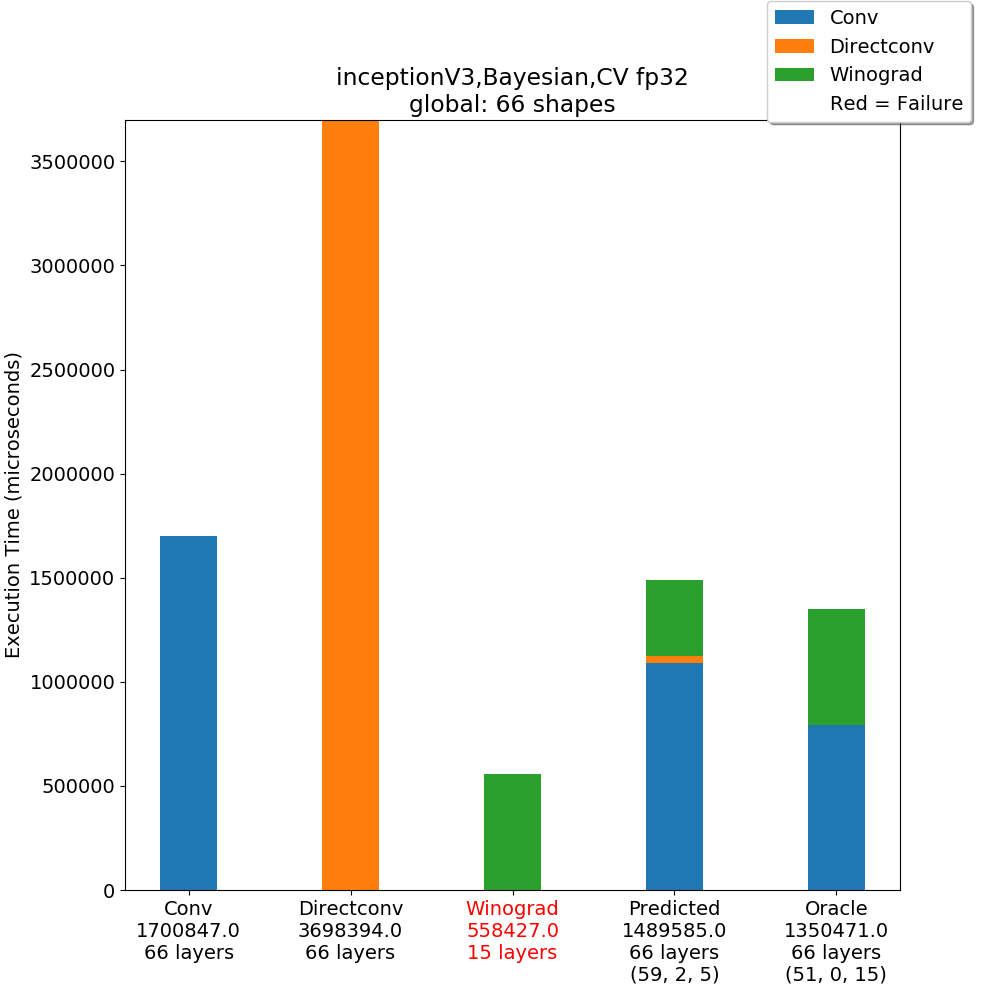}}
  \subfloat[All convolution layers Decision Tree]{\label{fig:inceptionGBDT}\includegraphics[width=0.55\textwidth]{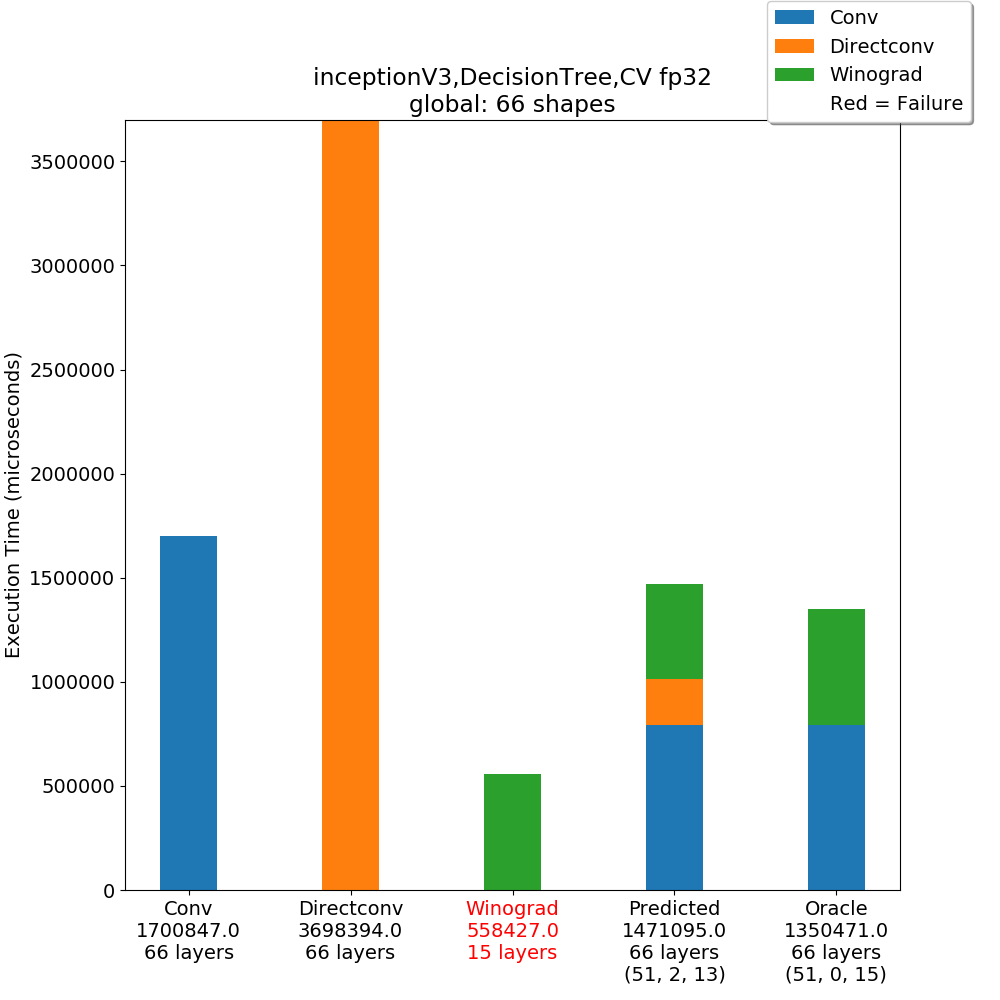}}
  \\
  \subfloat[Single layer naive Bayes]{\label{fig:inceptionSSBT}\includegraphics[width=0.55\textwidth]{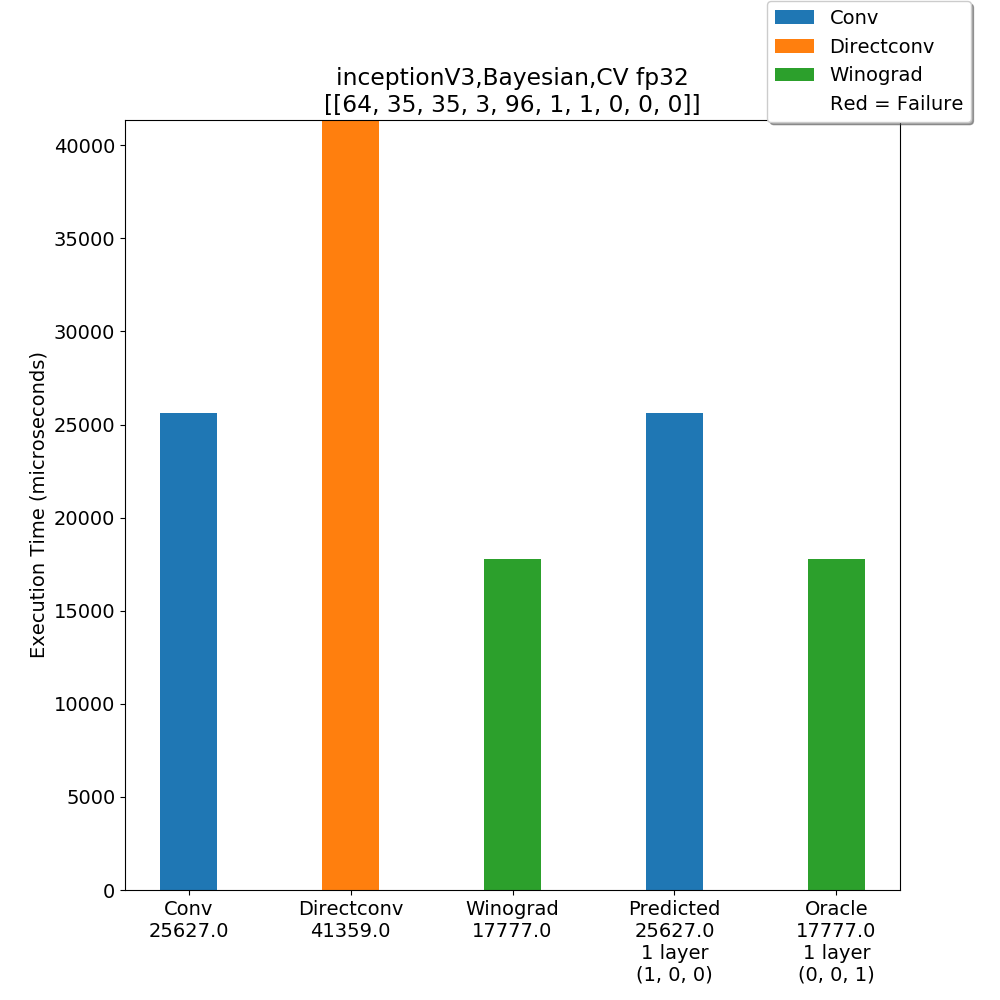}}
  \subfloat[Single layer Decision Tree]{\label{fig:inceptionSSDT}\includegraphics[width=0.55\textwidth]{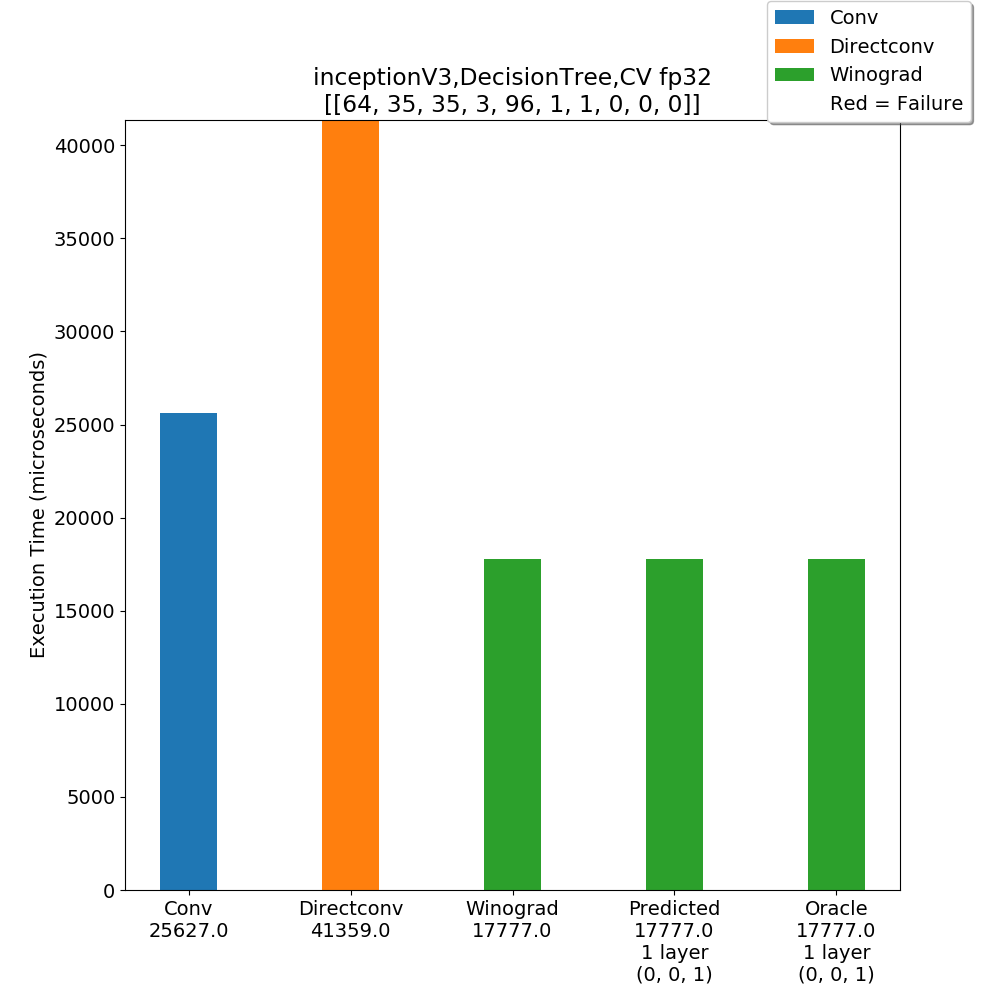}}
  \\

\end{figure}

\begin{figure}
  \centering
  \caption{MobileNets: naive Bayes vs Decision Tree}

  \subfloat[all shapes naive Bayes]{\label{fig:mobilenetsGBBT}\includegraphics[width=0.55\textwidth]{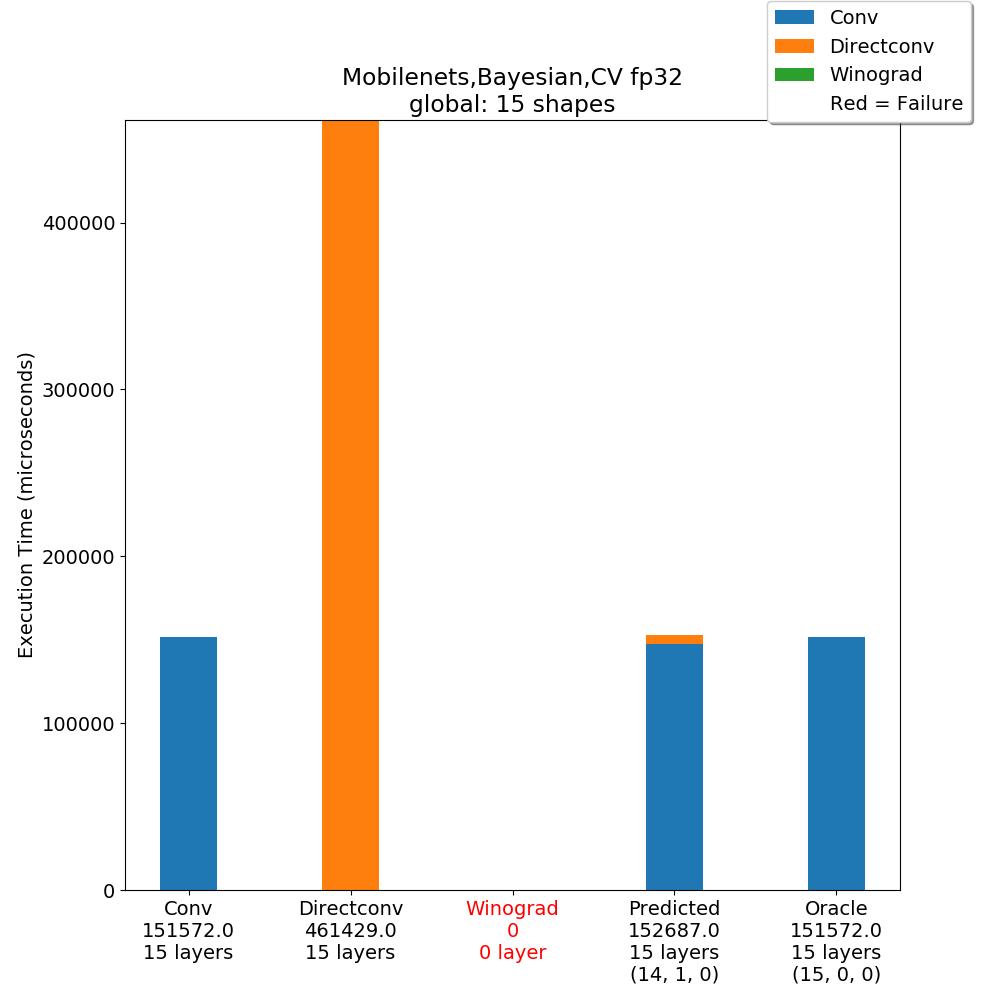}}
  \subfloat[all shapes Decision Tree]{\label{fig:mobilenetsGBDT}\includegraphics[width=0.55\textwidth]{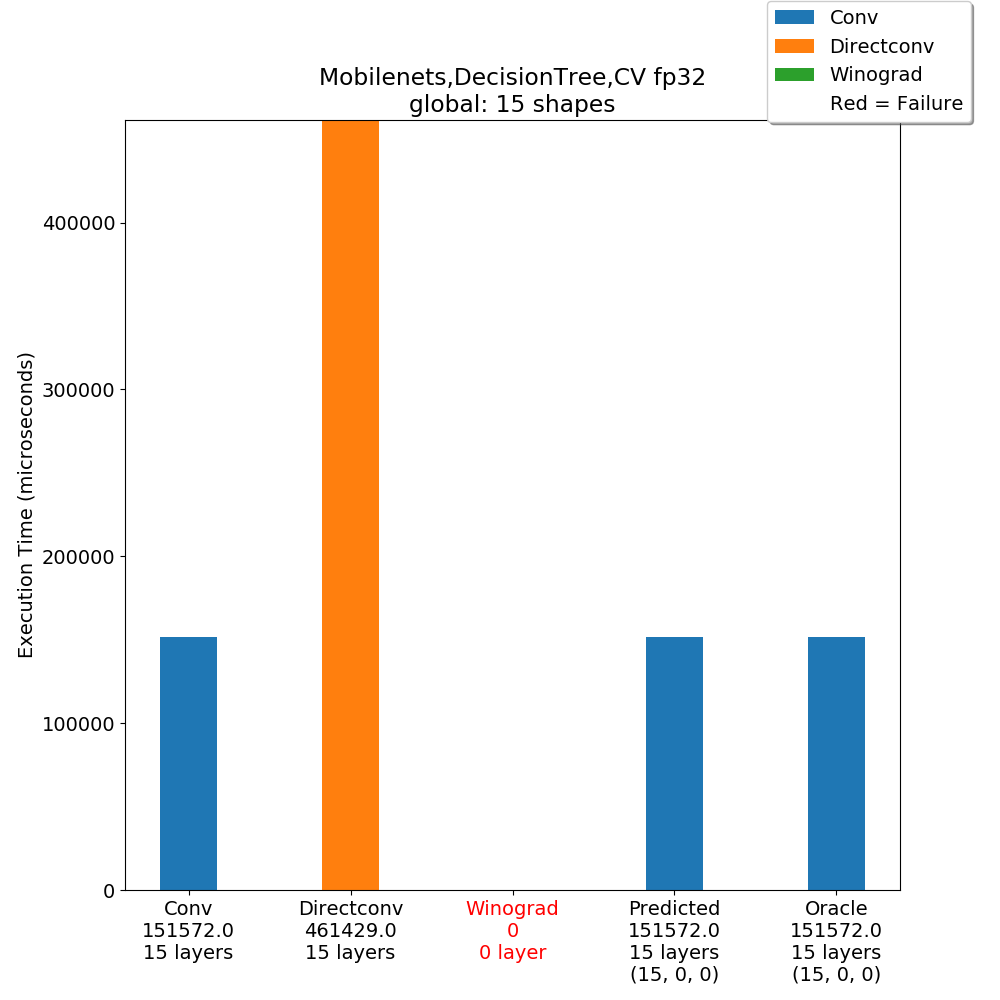}}
  \\
  \subfloat[single shape naive Bayes]{\label{fig:mobilenetsSSBT}\includegraphics[width=0.55\textwidth]{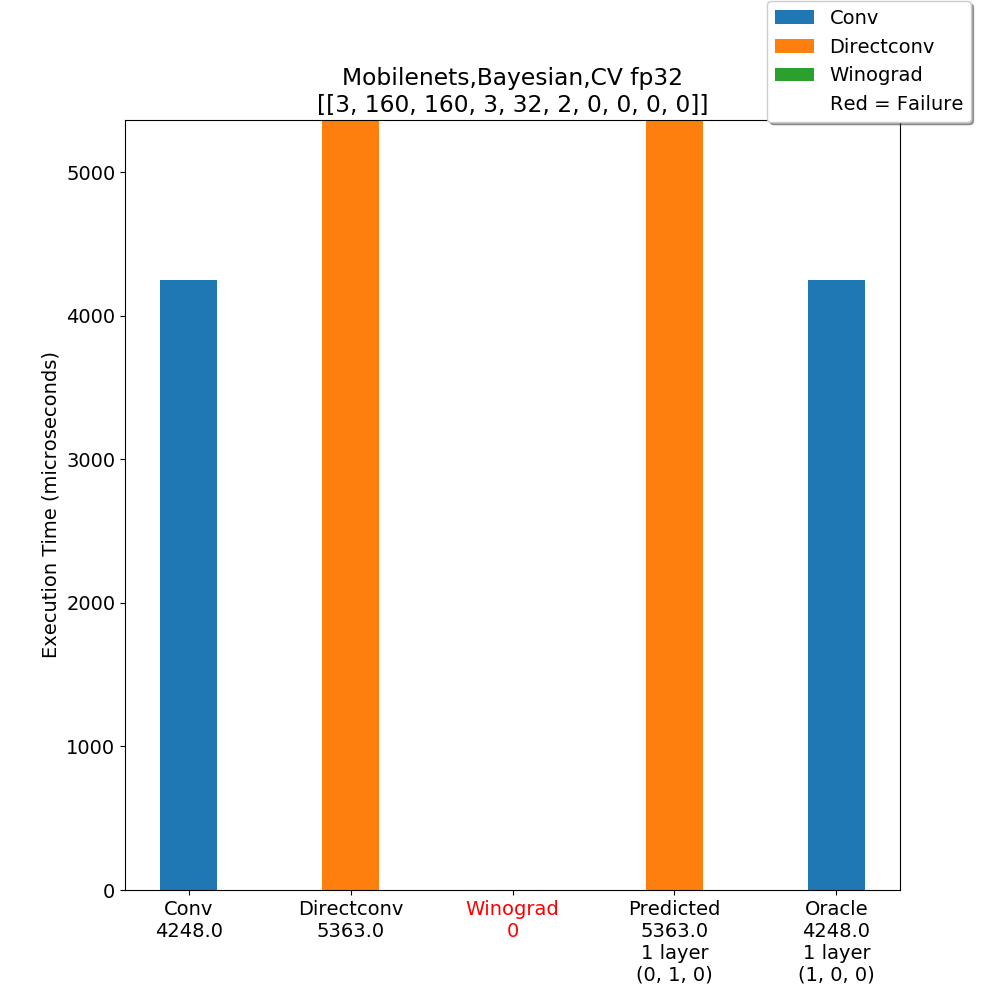}}
  \subfloat[single shape Decision Tree]{\label{fig:mobilenetsSSDT}\includegraphics[width=0.55\textwidth]{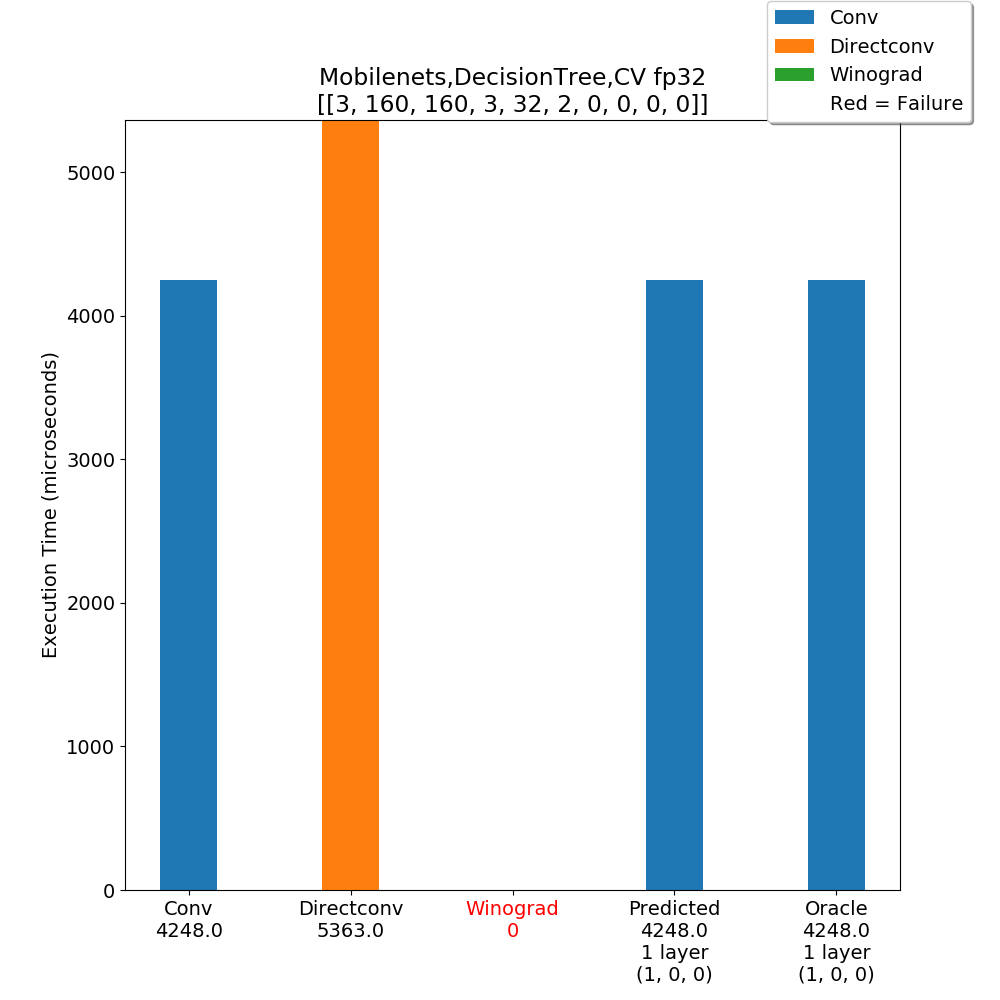}}
  \\

\end{figure}

In the present section we evaluate the performance of the predictive models trained by a DT and nBC against the implementations of the convolution operator provided by the ARM Compute Library. 
We analyze the accuracy of the classifiers as well as the execution time of the ARM Compute Library by using predictive models.
In Figure \ref{fig:inceptionGBDT} and Figure \ref{fig:mobilenetsGBDT}, we show the inference phase by using "Inception v3" CNN.

On top of each picture we indicate the classifiers used for training the predictive model, the numerical precision, the convolution neural network and the related number of layers.
In each figures, the Y axis represents the execution time needed to perform the convolution operator over of all the layers (microseconds).
The columns denote the execution time of each different implementation of convolution:
\begin{itemize}
    \item image to column and GEMM method;
    \item direct convolution method;
    \item Winograd method;
    \item method predicted by the model;
    \item the possible best algorithm Oracle. 
\end{itemize}

Below each column we report the total time in microseconds and the number of layers correctly completed.
In addition, the columns related to the predictive model and the oracle report a triple representing the times \textit{image to column and GEMM}, \textit{direct convolution} and \textit{Winograd} have been selected by that method.

Table \ref{tab:t1} and Table \ref{tab:t2} summarize the performance of the two models for both networks. 
In general, the predictive models perform better than each manually selected method except for one case, showing an high accuracy in the selection of the best implementation. 

Comparing the predictive models, that one based on the decision tree shows a better accuracy than nBC. However the overall computation times are still comparable.

The results for InceptionV3 are detailed in Figure \ref{fig:inceptionGBBT}. 
In the two top images the performance over the whole network are shown. 
The performances of the models learned from both classifiers are slightly worse than the optimal ones. However, the library that uses the model driven approach achieves an improvement over the handed-selected methods.\\ 
In the bottom images, a specific layer has been considered. In this example, the model based on the Bayesian approach erroneously selected a GEMM convolution instead of Winograd. Contrarily, the decision tree make the right prediction.

The same analysis is reported for MobileNets in Figure \ref{fig:mobilenetsGBBT}. For this network, the optimal convolution implementation was always GEMM.   
In general, the model learned by the decision tree shows better performance than the model based on naive Bayes classifier and the statically selected methods. 

\begin{table}[]
\caption {Classifier: naive Bayes classifier} \label{tab:t1} 
\centering
\resizebox{\textwidth}{!}{%
\begin{tabular}{ccccc}
\cline{2-5}
\multicolumn{1}{c|}{} & \multicolumn{1}{c|}{Accuracy} & \multicolumn{1}{c|}{vs IMG to column and GEMM} & \multicolumn{1}{c|}{vs Direct Convolution} & \multicolumn{1}{c|}{vs Winograd} \\ \hline
\multicolumn{1}{|c|}{MobileNets} & \multicolumn{1}{c|}{93.33\%} & \multicolumn{1}{c|}{0.99X} & \multicolumn{1}{c|}{3.02X} & \multicolumn{1}{c|}{winograd failed} \\ \hline
\multicolumn{1}{|c|}{Inception v3} & \multicolumn{1}{c|}{81.82\%} & \multicolumn{1}{c|}{1.14X} & \multicolumn{1}{c|}{2.48X} & \multicolumn{1}{c|}{winograd failed} \\ \hline
 &  &  &  &  \\ \hline
\end{tabular}%
}
\end{table}

\begin{table}[]
\caption {Classifier: Decision Tree} \label{tab:t2} 
\centering
\resizebox{\textwidth}{!}{%
\begin{tabular}{ccccc}
\cline{2-5}
\multicolumn{1}{c|}{} & \multicolumn{1}{c|}{Accuracy} & \multicolumn{1}{c|}{vs IMG to column and GEMM} & \multicolumn{1}{c|}{vs Direct Convolution} & \multicolumn{1}{c|}{vs Winograd} \\ \hline
\multicolumn{1}{|c|}{MobileNets} & \multicolumn{1}{c|}{100.00\%} & \multicolumn{1}{c|}{1.00X} & \multicolumn{1}{c|}{3.04X} & \multicolumn{1}{c|}{winograd failed} \\ \hline
\multicolumn{1}{|c|}{Inception v3} & \multicolumn{1}{c|}{96.97\%} & \multicolumn{1}{c|}{1.15X} & \multicolumn{1}{c|}{2.51X} & \multicolumn{1}{c|}{winograd failed} \\ \hline
 &  &  &  &  \\ \hline
\end{tabular}%
}
\end{table}

\section{Conclusions and future work}
\label{sec:conclusion}
We investigate the opportunity of using learned models to accelerate convolution operator in the case of a library that exhibits multiple implementations. 
We evaluate our predictive models on two different CNN, InsectionV3 and MobileNet (inference phase) on a low-power consumption ARM GPU. 

Our approach outperforms the ARM Compute Library with speed-ups up to $3x$. 
Future developments are going to focus on the improvement of the predictive models with more sophisticated and tunable classifiers. 
Also, since the dataset generation is the most expensive part of the proposed methodology, we are going to investigate solutions based on reinforcement learning. We will also explore predictive models for accelerating irregular application like graph analytics\cite{vella2018dynamic, bernaschi2018multilevel}, for selecting the best parallel strategy on GPU\cite{formisano2017accelerating} or optimizing communication on distributed systems\cite{di2017transparent}.

\section*{Acknowledgments}
We thank $Dividiti$ $Inc.$ for the huge support on CK and NNTest and for providing hardware resources.


\bibliographystyle {plain}
\bibliography {biblio}

\end{document}